\pdfoutput=1

\newcommand{\myheart}{\textsuperscript{$\heartsuit$}}
\newcommand{\myspadesuit}{\textsuperscript{$\spadesuit$}}
\newcommand{\mydiamondsuit}{\textsuperscript{$\diamondsuit$}}

\newcommand{\mysharp}{\textsuperscript{$\sharp$}}

\documentclass[11pt]{article}

\usepackage[final]{acl}

\usepackage{times}
\usepackage{latexsym}
\usepackage{amssymb}
\usepackage{amsmath}
\usepackage{colortbl}
\usepackage{booktabs} 
\usepackage{multirow}
\usepackage[T1]{fontenc}
\usepackage[most]{tcolorbox} 
\usepackage[utf8]{inputenc}

\usepackage{microtype}

\usepackage{inconsolata}

\usepackage{graphicx}

\usepackage{xcolor}

\definecolor{lightgreen}{RGB}{82, 184, 82}
\definecolor{lightblue}{RGB}{122, 163, 204}
\definecolor{lightorange}{RGB}{255,153,51}
\definecolor{periwinkle}{rgb}{0.8, 0.8, 1.0}

\title{\textit{CogSteer}: Cognition-Inspired Selective Layer Intervention for Efficiently Steering Large Language Models}

\author{\myspadesuit Xintong Wang, \myspadesuit Jingheng Pan, \mydiamondsuit Liang Ding
\\ \mysharp \textbf{Longyue Wang}\thanks{~~Corresponding Authors.}, \myspadesuit \textbf{Longqin Jiang}, \myheart \textbf{Xingshan Li} and \myspadesuit \textbf{Chris Biemann}$^*$ \\
        \myspadesuit Department of Informatics, Universität Hamburg,
        \mydiamondsuit The University of Sydney \\
        \mysharp Alibaba International Digital Commerce, \myheart Institute of Psychology, Chinese Academy of Sciences \\
        {\tt\small \myspadesuit\{xintong.wang, jingheng.pan, longqin.jiang, chris.biemann\}@uni-hamburg.de} \\
        {\tt\small \mydiamondsuit liangding.liam@gmail.com, \mysharp wanglongyue.wly@alibaba-inc.com, \myheart lixs@psych.ac.cn}
        }

\begin{document}
\maketitle
\begin{abstract}
Large Language Models (LLMs) achieve remarkable performance through pretraining on extensive data. This enables efficient adaptation to diverse downstream tasks. However, the lack of interpretability in their underlying mechanisms limits the ability to effectively steer LLMs for specific applications. In this work, we investigate the intrinsic mechanisms of LLMs from a cognitive perspective using eye movement measures. Specifically, we analyze the layer-wise correlation between human cognitive indicators and LLM representations. Building on these insights, we propose a heuristic approach for selecting the optimal steering layer to modulate LLM semantics. To this end, we introduce an efficient selective layer intervention based on prominent \textit{parameter-efficient fine-tuning} methods, which conventionally adjust either all layers or only the final layer. Additionally, we present an \textit{implicit layer contrastive intervention} during inference to steer LLMs away from toxic outputs. Extensive experiments on natural language understanding, reasoning, and generation tasks, conducted on GPT-2, Llama2-7B, and Mistral-7B, demonstrate the effectiveness and efficiency of our approach. As a model-agnostic framework, it enhances the interpretability of LLMs while improving efficiency for safe deployment.
\end{abstract}

\section{Introduction}

\begin{figure}[htbp]
  \centering
  \includegraphics[width=0.9\columnwidth]{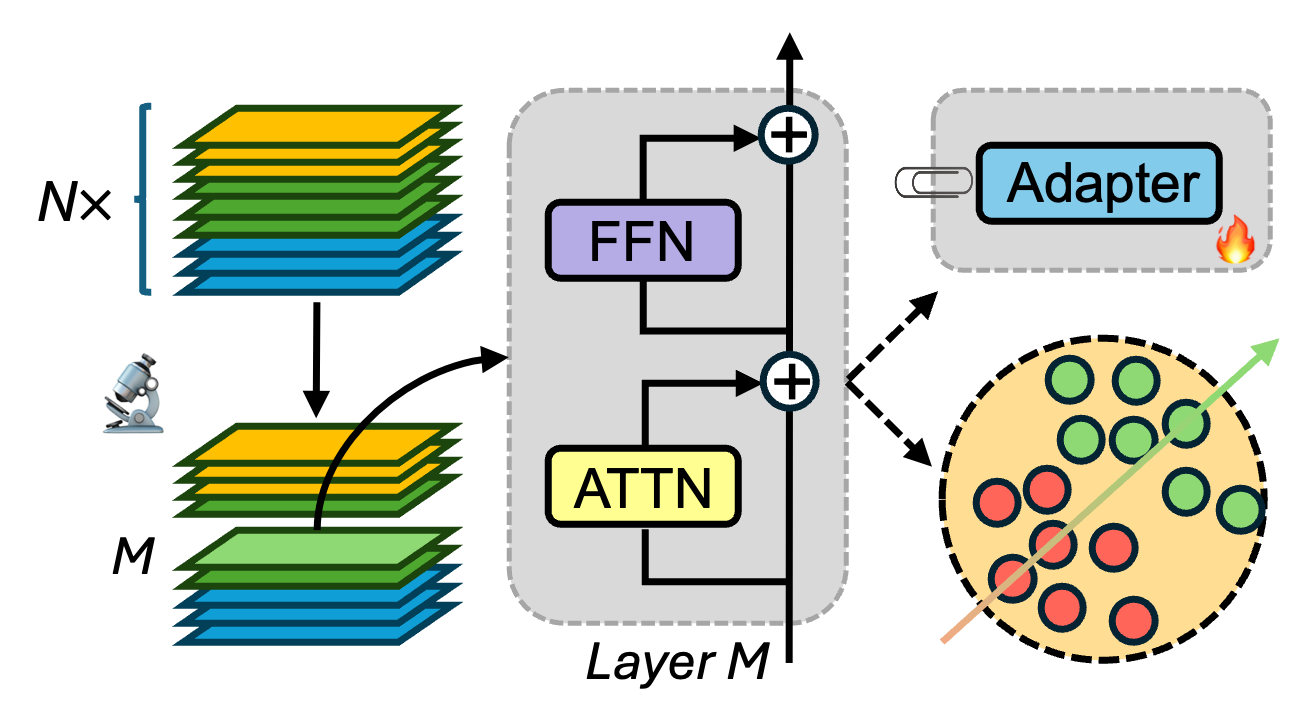}
  \caption{\textbf{Demonstration of CogSteer Intervention}. For an \textit{N}-layer LLM, we first heuristically find the optimal layer \textit{M} for semantic intervention. The upper block represents an adapter that is fine-tuned and inserted into the frozen layer \textit{M}. The bottom block illustrates the operation of the attention module in \textit{M} to steer the semantic direction towards safer outputs during inference.}
  \vspace{-10pt}
  \label{fig:main} 
\end{figure}

Large Language Models (LLMs) \cite{dubey2024llama, yang2024qwen2, guo2025deepseek} have demonstrated strong capabilities in natural language understanding and reasoning \cite{zhao2024marcoo1openreasoningmodels, zhao2023survey, wei2022emergent, yin2025towards, zeng2025marco} through pretraining on large datasets, followed by instruction tuning and alignment with human values \cite{wei2021finetuned, ouyang2022training}. Consequently, LLMs achieve excellent performance on downstream tasks with fine-tuning. However, their lack of interpretability and transparency limits the development of efficient fine-tuning and inference methods.


To understand intrinsic mechanisms of LLMs, previous work has introduced various interpretability methods, including training linear classifiers as probes on top of hidden representations~\cite{belinkov2022probing}, projecting representations into vocabularies~\cite{geva2022transformer}, and intervening in the computation path, such as knowledge neurons~\cite{dai2022knowledge} and circuits~\cite{conmy2023towards,ghandehariounpatchscopes}. However, these methods, which focus on a limited set of predefined classes, concepts, or prompts, have practical limitations in terms of scalability and generalization.

In this work, we introduce a novel interpretability analysis method that leverages eye movement data~\cite{luke2018provo, hollenstein2020zuco, colman2022geco} collected by cognition researchers to study human reading behaviors. Through correlation experiments, we find that LLM hidden states exhibit a strong correlation with human gaze, peaking in the middle layers. Using eye movement measures such as \textit{fixation} and \textit{regression} \cite{rayner1998eye} as human-interpretable indicators, we observe a hierarchical progression in LLMs, from initial syntactic and semantic processing to deeper integration and final prediction. Additionally, a comparison of correlation results between natural reading and task-specific reading suggests that the upper layers of LLMs are more capable of reasoning. Furthermore, advanced LLMs such as Llama, which incorporate instruction tuning and reinforcement learning from human feedback (RLHF), demonstrate enhanced reasoning capabilities, even in the middle layers.

LLM intervention \cite{poth2023adapters, dong2024survey} refers to fine-tuning or applying inference methods to steer their semantics to align with specific tasks and data distributions. In addition to enhancing the understanding of LLM behavior, our interpretability analysis reveals that different layers serve distinct functions, with middle layers playing a crucial role in deeper syntactic and semantic processing. This enables us to first identify the most suitable layer for intervention, thereby improving task-specific performance. Based on these insights, we propose a heuristic approach for selecting the optimal steering layer for semantic intervention.

To achieve this, we refine prominent Parameter-Efficient Fine-Tuning (PEFT) methods, which traditionally adjust either all layers or only the last layer. As shown in Figure~\ref{fig:main}, our proposed \textit{CogSteer} framework first identifies the most suitable layer $M$ for semantic intervention based on the task. Instead of fine-tuning all layers or only the last layer, our method fine-tunes only the selected layer, enabling LLMs to better adapt to specific tasks and datasets. The number of learnable parameters in \textit{CogSteer} is significantly reduced, requiring only $1/N$ of the parameters in LLMs, thereby improving parameter efficiency. Furthermore, we propose an \textit{implicit layer contrastive intervention} method during inference, which efficiently identifies and steers semantics toward safer generation directions to evaluate the effectiveness of our proposed selective layer intervention.

Through extensive evaluations across diverse tasks and datasets, we demonstrate that the proposed selective layer intervention method achieves comparable or even superior performance with fewer parameters compared to the full-layer intervention baseline. Specifically, we observed an average absolute improvement of \textbf{+1.7} on the GLUE benchmark for Llama2-7B, and an average absolute improvement of \textbf{+5.8} on the GLUE benchmark for Mistral-7B, with only \textbf{3.1\%} of the parameters involved in full-layer fine-tuning. Moreover, in experiments on generation tasks, language toxification \cite{dementieva2025multilingual}, and detoxification \cite{leong2023self}, our method achieves a \textbf{+1.85\%} improvement in toxification compared to full-layer intervention and a \textbf{+13.45\%} improvement in detoxification as compared to last-layer intervention. 

Our \textbf{main contributions} are as follows:

(1) We are the first to propose leveraging eye movement measures to analyze the layer-wise behavior of LLMs. We publicly release the probing code~\footnote{https://github.com/Ethanscuter/CogSteer} to facilitate further research on interpretability from a cognitive perspective.
(2) Through correlation analysis, we demonstrate a hierarchical progression in LLMs and introduce a heuristic steering layer selection method for efficient layer intervention.
(3) Extensive experiments validate the effectiveness of our proposed method across various language understanding, reasoning, and generation tasks, contributing to the development of efficient and explainable foundation models.

\begin{figure*}[htbp] 
  \centering
  \includegraphics[width=1\textwidth]{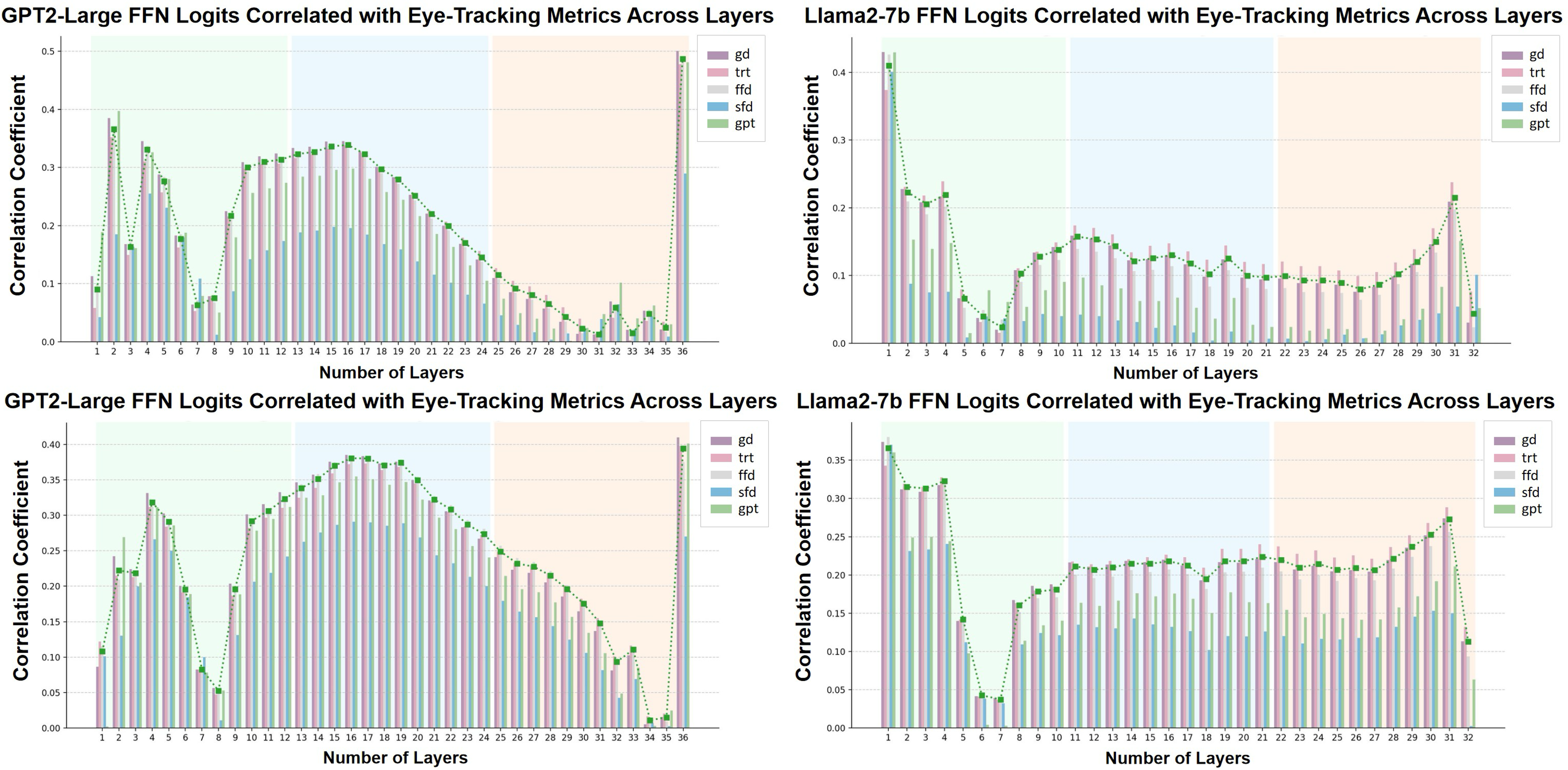} 
  \caption{\textbf{Correlation Results Comparison: \textit{Natural Reading (NR) vs.\ Task-Specific Reading (TSR)}.} Correlation results are shown for GPT-2 Large NR (36 layers, top-left), GPT-2 Large TSR (36 layers, bottom-left), Llama2-7B NR (32 layers, top-right), and Llama2-7B TSR (36 layers, bottom-right). Green, blue, and orange boxes indicate the premature, middle, and mature buckets, respectively.}
  \vspace{-10pt}
  \label{fig:nrvstsk} 
\end{figure*}

\section{Related Work}

\paragraph{Interpretability} research is essential for uncovering the mechanisms of LLMs and ensuring their safe and trustworthy deployment as foundation models. Various studies have analyzed how knowledge is stored in LLMs \cite{goldowsky2023localizing, stolfo-etal-2023-mechanistic, rai2024practical, bills2023language, geva2022transformer}, focusing on concepts such as knowledge neurons \cite{dai2022knowledge} and circuits \cite{conmy2023towards, yao2024knowledge}. Moreover, \citet{wang-etal-2024-probing} reveal that GPT-2 predicts tokens more similarly to humans than shallow language models but lacks engagement with LLMs. \citet{oh-schuler-2023-transformer} analyzes how model scale and training data influence surprisal-based predictions of reading time. \citet{gao-etal-2023-roles} investigates the alignment between model and human attention. Unlike previous works, our method investigates the interpretability of LLMs through human-interpretable indicators based on eye movement theory, enabling a more precise understanding and control of model behavior (see §~\ref{sec:explain} and ~\ref{sec:heuristic}).

\paragraph{Parameter-Efficient Semantic Steering} PEFT \cite{han2024parameter,wang-etal-2023-using}, including Adapter \cite{houlsby2019parameter,poth2023adapters} and LoRA \cite{hulora}, has gained popularity due to its ability to maintain a large number of frozen parameters in LLMs for generality while introducing only a small number of trainable parameters per task. Our method for semantic steering via fine-tuning enhances PEFT methods by incorporating a selective layer intervention strategy, reducing computational costs while achieving superior performance (see §~\ref{sec:sft}). Furthermore, steering semantics during inference offers an even more efficient approach. Contrastive decoding \cite{li2022contrastive, sennrich2023mitigating, wang-etal-2024-mitigating} guides the generation process by comparing two output distributions. In contrast, our proposed implicit layer contrastive intervention efficiently identifies and steers the semantics of LLMs toward safe directions during inference (see §~\ref{sec:inference}).

\section{From Human Gaze to LLM Behavior}
\label{sec:gaze}

Recent studies \cite{geva2021transformer, schuster2022confident} suggest that feed-forward networks (FFNs) function similarly to neural memory networks, capturing syntactic and semantic features as well as factual knowledge \cite{chuang2023dola, zhang2024comprehensive}. Meanwhile, research in cognitive science \cite{rayner1998eye} has shown that eye movement measures provide insights into the time required for human readers to process syntax, semantics, and integrate information. Motivated by these findings, we leverage eye movement measures to analyze their correlation with the hidden states of FFNs across different layers of LLMs.

\textbf{Models.} Correlation studies are conducted on the GPT-2 model \cite{radford2019language} at different sizes (12-layer small, 24-layer medium, 36-layer large) and the 32-layer Llama2-7B \cite{touvron2023llama}. Both GPT-2 and Llama2-7B use a decoder-only transformer architecture. Comparing earlier LLMs, such as GPT-2, with more advanced models like Llama2 provides valuable insights into their similarities and differences while also enhancing the robustness and generalizability of our findings.

\textbf{Eye-movement Experiments and Datasets.} We conduct a correlation analysis under two experimental conditions: natural reading and task-specific reading. For natural reading, we use the Provo \cite{luke2018provo}, GECO \cite{colman2022geco}, and ZuCo 2.0 \cite{hollenstein2020zuco} datasets. The ZuCo 2.0 dataset also includes task-specific reading experiments. In task-specific reading, participants are required to determine whether a specific relation type is present in a sentence. Relation detection is a high-level semantic and reasoning task that involves complex cognitive processing.

\textbf{Correlation Analysis.} Let $S_{j}$ denote the $j$-th sentence, consisting of $n_{j}$ words \( w_1, w_2, \dots, w_{n_j} \). For each word \( w_i \) in sentence \( S_j \), we consider five eye movement measures: $e_i^{(k)}, k \in \{\mathrm{\textit{sfd}},\, \mathrm{\textit{ffd}},\, \mathrm{\textit{gd}},\, \mathrm{\textit{trt}},\, \mathrm{\textit{gpt}}\}$, where each measure represents a scalar value. The hidden state at layer \( l \) of the LLM for word \( w_i \) is denoted as $\mathbf{h}_{l, i} \in \mathbb{R}^d$, where \( d \) is the dimensionality of the hidden states.

To analyze the relationship between eye movement measures and LLM hidden states, we compute the \textit{Pearson correlation} between each eye movement measure and the corresponding hidden states at each layer. Specifically, we concatenate hidden states and eye movement measures across all words in the dataset and apply \textit{principal component analysis} to obtain a scalar representation of hidden states, ensuring alignment with eye movement measures (see Appendix~\ref{sec:analysis_B} for details). The correlation is then defined as:

{\small
\begin{align}
\rho_{l,k} = 
\frac{\sum\limits_{i=1}^{n_{\textit{total}}} \bigl( h_{l,i} - \bar{h}_l \bigr)\bigl( e_i^{(k)} - \bar{e}^{(k)} \bigr)}
{\sqrt{\sum\limits_{i=1}^{n_{\textit{total}}} \bigl( h_{l,i} - \bar{h}_l \bigr)^2}
\sqrt{\sum\limits_{i=1}^{n_{\textit{total}}} \bigl( e_i^{(k)} - \bar{e}^{(k)} \bigr)^2}},
\label{eq:correlationequ}
\end{align}
}where \( h_{l,i} \) represents the processed hidden states aligned with the eye movement measures, and \( \bar{h}_l \) and \( \bar{e}^{(k)} \) are their respective means.

\textbf{Results and Finding 1.} Based on the correlation calculation in Eq.~\ref{eq:correlationequ}, we first analyze the correlation results for the natural reading task. To facilitate interpretation, we divide the layers of both models into three equal groups: \textcolor{lightgreen}{\textit{premature}}, \textcolor{lightblue}{\textit{middle}}, and \textcolor{lightorange}{\textit{mature}}. The tripartite division of layers into premature, middle, and mature buckets is motivated by recent interpretability studies \cite{geva2021transformer, schuster2022confident, chuang2023dola, zhang2024comprehensive}, which consistently observe a coarse-grained functional structure in LLMs: lower layers predominantly capture syntactic patterns, middle layers encode semantic and contextual information, and upper layers are often involved in reasoning and factual retrieval. Figures~\ref{fig:nrvstsk} (\textit{upper}) and ~\ref{fig:correlation} present the correlation results between various eye movement measures and the hidden states of the LLMs, illustrating how these values evolve across layers.

The results indicate that the hidden states of different LLM layers exhibit a clear and strong correlation with human gaze, peaking in the middle bucket and reaching a secondary peak in the mature bucket. This trend is consistent across different eye movement measures and LLMs with varying layer sizes. Considering the nature of these eye movement measures, the increase in correlation in the \textcolor{lightgreen}{\textit{premature bucket}} suggests that LLMs begin processing tokens by integrating syntactic and semantic features, reflecting an initial focus on token processing. In the \textcolor{lightblue}{\textit{middle bucket}}, the further increase in correlation signifies deeper syntactic and semantic processing, with the peak indicating the integration of linguistic features. In the \textcolor{lightorange}{\textit{mature bucket}}, the secondary peak likely reflects the final integration of information for word prediction.

The overall trend across the three buckets is similar for GPT-2 and Llama2-7B. For a detailed analysis of eye movement measures and additional findings, please refer to Appendix~\ref{sec:analysis_B}.



\label{sec:explain}
\begin{tcolorbox}[
    enhanced,
    colback=white,                    
    colframe=black,                   
    colbacktitle=blue!50!white,      
    coltitle=white,                   
    title=Finding 1. (Layer-wise functionality),
    fonttitle=\bfseries,              
    boxrule=0.5pt,                    
    arc=2pt,                          
    top=4pt, bottom=4pt, left=6pt, right=6pt,  
    halign title=left,              
]

LLM hidden states exhibit a strong correlation with human gaze, characterized by three distinct rises across layers. This pattern suggests a hierarchical progression from initial syntactic and semantic processing to deeper integration and final prediction.

\end{tcolorbox}

\textbf{Results and Finding 2.} Empirically, LLMs are trained with the next-token prediction objective. Modern LLMs demonstrate strong language understanding and reasoning abilities, raising the question: \textit{Are LLMs merely next-token predictors, or are they task reasoners?} Figure~\ref{fig:nrvstsk} presents a comparison between natural reading and task-specific reading. The correlation patterns suggest that LLMs function as both next-token predictors and reasoners, as the trends in task-specific reading closely resemble those observed in natural reading.


Notably, for both GPT-2 and Llama2-7B models, the correlation values in the \textcolor{lightblue}{\textit{middle bucket}} during task-specific reading are higher than those observed during natural reading. In particular, for the Llama2-7B model, these values remain consistently higher in task-specific reading. We hypothesize that this indicates Llama2-7B is better suited for reasoning tasks and that the layers in the \textcolor{lightblue}{\textit{middle bucket}} and \textcolor{lightorange}{\textit{mature bucket}} are activated when processing complex tasks, as it is trained on a larger text corpus and incorporates more advanced post-training techniques.

\begin{tcolorbox}[
    enhanced,
    colback=white,                    
    colframe=black,                   
    colbacktitle=blue!50!white,      
    coltitle=white,                   
    title=Finding 2. (LLM functions as both next-token predictor and task reasoner),
    fonttitle=\bfseries,              
    boxrule=0.5pt,                    
    arc=2pt,                          
    top=4pt, bottom=4pt, left=6pt, right=6pt,  
    halign title=left,              
]

LLMs function as both next-token predictors and reasoners, with overall correlation trends aligning with human cognition indicators. Advanced training methods enhance their ability to reason and handle complex tasks.

\end{tcolorbox}

\section{Method}
\subsection{Heuristic Steering Layer Selection}
\label{sec:heuristic}

A better understanding of LLM mechanisms will help in precisely and efficiently controlling their behaviors, particularly for semantic steering. We argue that the predominant parameter-efficient fine-tuning (PEFT) methods \cite{han2024parameter}, which by default intervene in the last layer or across all layers, are not optimal\footnote{Recent work~\cite{yu2023promptst} on fine-tuning speech translation models also supports our hypothesis.}. Instead, we propose an efficient heuristic steering layer selection strategy for intervention, based on our cognition-inspired interpretability analysis detailed in Section~\ref{sec:explain}.

For semantic steering in LLMs, the layers in the middle bucket are the most suitable candidates for intervention. These layers handle further token processing, information integration, and preliminary reasoning. Additionally, the residual connections \cite{he2016deep} in transformer layers allow the semantic intervention to flow and evolve gradually, avoiding abrupt changes in the final prediction \cite{chuang2023dola}.

From a task-oriented perspective, we apply PEFT methods and inference-only methods to the candidate layers in the middle bucket, using a small portion of data, like the validation set, to search for and select the best-performing layer that suits the task scenario. Formally, given \( M \) as the best layer for intervention, \( J \) represents a set of candidate layers in the middle bucket \( \left( \frac{N}{3} \leq M \leq \frac{2N}{3} \right) \). \( D \) denotes the validation set. We search for the layer \( M' \) that yields the best task score or loss performance, as follows:

{\small
\begin{align}
M' = \arg \max_{l \in M} Score\left(D ; P\left(\cdot \mid x_t,l\right)\right).
\label{eq:M}
\end{align}
}

Later, we will demonstrate the effectiveness of the heuristic steering layer selection approach in language understanding, reasoning, and generation tasks, as discussed in Section \ref{sec:exp}.


\subsection{Layer Intervention via Fine-tuning}
\label{sec:sft}
PEFT approaches, such as additive fine-tuning (i.e., adapters) and reparameterized fine-tuning (i.e., LoRA), are among the most popular due to their efficiency, as they require only a small set of new parameters for task-specific fine-tuning. Let the parameters of an LLM consist of a set of pre-trained, frozen parameters \( \phi(\cdot) \) and a set of newly introduced parameters in the inserted block \( \psi(\cdot) \). Our layer intervention via fine-tuning to steer semantics in LLM and predicts the next token as follows:

{\scriptsize
\begin{align}
y(x_t) 
= \operatorname{softmax}\bigl(
  \operatorname{logit}_{\phi, \psi}\bigl(
    FFN_\phi^N(x_t) \mid F_{\phi, \psi}^M(x_t), y_{<t}
  \bigr)
\bigr).
\end{align}
}Here, \(FFN_{\phi}^N(x_t) \) represents the hidden state of the FFN in the final layer with frozen parameters \( \phi(\cdot) \), used for token prediction over a vocabulary. \( M \) denotes the best layer for semantic intervention, determined by Equation~\ref{eq:M}. \( F_{\phi, \psi}^M(x_t) \) indicates that the fusion layer in the LLM integrates new parameters \( \psi(\cdot) \) from the newly added block, aligned with the frozen parameters \( \phi(\cdot) \).

Our cognitive-inspired selective layer intervention method is an adaptive fine-tuning strategy that identifies the best layer for both effective semantic steering and task performance. Moreover, as our method only operates on a single layer rather than all layers, it significantly reduces computational resources and time, while also avoiding catastrophic forgetting \cite{luo2023empirical,li2024revisiting}.

\begin{table*}[!h]
\centering
    \small 
    \setlength\tabcolsep{6pt} 
    \renewcommand{\arraystretch}{1.2} 

\definecolor{mygray}{gray}{0.55}
\begin{tabular}{lccccccccc}
\toprule
\multirow{2}{*}{\textbf{Model}} & \multicolumn{8}{c}{\textbf{VAL-SET}} \\
\cmidrule(lr){2-9}
& \textbf{MNLI-M} & \textbf{MNLI-MM} & \textbf{MRPC} & \textbf{QNLI} & \textbf{QQP} & \textbf{RTE} & \textbf{SST-2} & \textbf{WNLI} & \textbf{Avg.} \\
\midrule
\multirow{2}{*}{GPT2-L} & 
        \cellcolor{blue!20} 78.0 \textcolor{mygray}{\textsubscript{\textit{l-}19}} & \cellcolor{blue!20} 79.5 \textcolor{mygray}{\textsubscript{\textit{l-}19}} & \cellcolor{green!20} 85.5 \textcolor{mygray}{\textsubscript{\textit{l-}20}} & \cellcolor{orange!20} 83.3 \textcolor{mygray}{\textsubscript{\textit{l-}19}} & \cellcolor{orange!20} 80.6 \textcolor{mygray}{\textsubscript{\textit{l-}19}} & \cellcolor{green!20} 71.1 \textcolor{mygray}{\textsubscript{\textit{l-}19}} & 
        \cellcolor{orange!20} 92.9 \textcolor{mygray}{\textsubscript{\textit{l-}19}} & \cellcolor{green!20} 53.5 \textcolor{mygray}{\textsubscript{\textit{l-}19}} & \cellcolor{orange!20}78.1\\
         & 82.1 & 83.5 & 83.1 & 85.2 & 82.6 & 70.4 & 93.6 & 53.5 & 79.2 \\
        
        \multirow{2}{*}{Llama2-7B} & 
        \cellcolor{blue!20} 86.4 \textcolor{mygray}{\textsubscript{\textit{l-}14}} & \cellcolor{blue!20} 87.1 \textcolor{mygray}{\textsubscript{\textit{l-}14}} & \cellcolor{green!20} 86.5 \textcolor{mygray}{\textsubscript{\textit{l-}14}} & \cellcolor{blue!20} 89.3 \textcolor{mygray}{\textsubscript{\textit{l-}14}} & \cellcolor{blue!20} 83.3 \textcolor{mygray}{\textsubscript{\textit{l-}14}} & \cellcolor{green!20} 75.5 \textcolor{mygray}{\textsubscript{\textit{l-}14}} & \cellcolor{orange!20} 95.8 \textcolor{mygray}{\textsubscript{\textit{l-}14}} & \cellcolor{green!20} 56.4 \textcolor{mygray}{\textsubscript{\textit{l-}19}} & \cellcolor{green!20}82.5\\
        & 89.0 & 89.3 & 86.3 & 91.9 & 85.6 & 65.3 & 96.7 & 56.3 & 82.5 \\
        
        \multirow{2}{*}{Mistral-7B} & 
        \cellcolor{blue!20} 87.3 \textcolor{mygray}{\textsubscript{\textit{l-}12}} & \cellcolor{orange!20} 88.1 \textcolor{mygray}{\textsubscript{\textit{l-}12}} & \cellcolor{green!20} 86.9 \textcolor{mygray}{\textsubscript{\textit{l-}12}} & 
        \cellcolor{green!20} 91.4 \textcolor{mygray}{\textsubscript{\textit{l-}14}} & \cellcolor{green!20} 84.6 \textcolor{mygray}{\textsubscript{\textit{l-}12}} & \cellcolor{green!20} 80.1 \textcolor{mygray}{\textsubscript{\textit{l-}12}} & 
        \cellcolor{orange!20} 95.8 \textcolor{mygray}{\textsubscript{\textit{l-}14}} & \cellcolor{green!20} 56.3 \textcolor{mygray}{\textsubscript{\textit{l-}12}} & \cellcolor{green!20}83.8\\
        & 89.5 & 89.7 & 82.2 & 81.7 & 78.1 & 58.9 & 96.7 & 56.3 & 79.1 \\
        \hline
        \hline
        \multirow{2}{*}{\textbf{Model}} & \multicolumn{8}{c}{\textbf{TEST-SET}} \\
        \cmidrule(lr){2-9}
        & \textbf{MNLI-M} & \textbf{MNLI-MM} & \textbf{MRPC} & \textbf{QNLI} & \textbf{QQP} & \textbf{RTE} & \textbf{SST-2} & \textbf{WNLI} & \textbf{Avg.} \\
        \midrule
        \multirow{2}{*}{GPT2-L} & 
        \cellcolor{blue!20} 79.3 \textcolor{mygray}{\textsubscript{\textit{l-}19}} & \cellcolor{blue!20} 79.3 \textcolor{mygray}{\textsubscript{\textit{l-}19}} & \cellcolor{green!20} 83.0  \textcolor{mygray}{\textsubscript{\textit{l-}20}} & \cellcolor{orange!20}84.1 \textcolor{mygray}{\textsubscript{\textit{l-}19}} & \cellcolor{green!20} 65.6 \textcolor{mygray}{\textsubscript{\textit{l-}19}} & 
        \cellcolor{green!20} 64.6 \textcolor{mygray}{\textsubscript{\textit{l-}19}} & \cellcolor{orange!20} 92.4 \textcolor{mygray}{\textsubscript{\textit{l-}19}} & \cellcolor{blue!20} 58.9 \textcolor{mygray}{\textsubscript{\textit{l-}19}} & \cellcolor{orange!20}75.8\\
        & 82.6 & 83.0 & 82.7 & 85.6 & 65.6 & 62.6 & 93.5 & 61.6 & 77.1 \\

        \multirow{2}{*}{Llama2-7B} & 
        \cellcolor{blue!20} 82.9 \textcolor{mygray}{\textsubscript{\textit{l-}14}} & \cellcolor{blue!20} 86.3 \textcolor{mygray}{\textsubscript{\textit{l-}14}} & \cellcolor{green!20} 83.4 \textcolor{mygray}{\textsubscript{\textit{l-}14}} & \cellcolor{blue!20} 88.5 \textcolor{mygray}{\textsubscript{\textit{l-}14}} & \cellcolor{blue!20} 68.8 \textcolor{mygray}{\textsubscript{\textit{l-}14}} & \cellcolor{green!20} 74.7 \textcolor{mygray}{\textsubscript{\textit{l-}14}} & \cellcolor{green!20} 95.2 \textcolor{mygray}{\textsubscript{\textit{l-}14}} & \cellcolor{green!20} 64.4 \textcolor{mygray}{\textsubscript{\textit{l-}19}} & \cellcolor{green!20}80.5\\
        & 89.5 & 88.8 & 80.5 & 92.1 & 71.6 & 58.2 & 93.5 & 55.5 & 78.7 \\

        \multirow{2}{*}{Mistral-7B} & 
        \cellcolor{blue!20} 87.1 \textcolor{mygray}{\textsubscript{\textit{l-}12}} & \cellcolor{orange!20} 87.5 \textcolor{mygray}{\textsubscript{\textit{l-}12}} & \cellcolor{green!20} 86.6 \textcolor{mygray}{\textsubscript{\textit{l-}12}} & \cellcolor{green!20} 91.7 \textcolor{mygray}{\textsubscript{\textit{l-}14}} & \cellcolor{green!20} 70.5 \textcolor{mygray}{\textsubscript{\textit{l-}12}} & \cellcolor{green!20} 81.0 \textcolor{mygray}{\textsubscript{\textit{l-}12}} & 
        \cellcolor{orange!20} 95.9 \textcolor{mygray}{\textsubscript{\textit{l-}14}} & \cellcolor{green!20} 65.8 \textcolor{mygray}{\textsubscript{\textit{l-}12}} & \cellcolor{green!20}83.2\\
        & 89.7 & 89.4 & 80.4 & 81.3 & 62.7 & 52.3 & 97.3 & 65.1 & 77.3 \\
\bottomrule
\end{tabular}
\caption{\textbf{Evaluation on GLUE Benchmark.} MRPC and QQP are reported using F1, while the other tasks are reported using Accuracy. A \textcolor{lightgreen}{green box} indicates that the single-layer intervention outperforms the full-layer intervention, an \textcolor{lightorange}{orange box} denotes comparable performance, and a \textcolor{periwinkle}{blue box} indicates slightly lower performance.}
\vspace{-10pt}
\label{tab:glue}
\end{table*}

\subsection{Layer Intervention during Inference}
\label{sec:inference}

Efficient semantic steering can be achieved via fine-tuning. However, an even more efficient approach is to steer the semantics of LLMs during inference without introducing additional parameters. Motivated by \citet{li2022contrastive, leong2023self}, which contrast outputs from either a less capable model or outputs induced by a negative prompt, we propose an \textit{implicit layer contrastive intervention} method during inference. First, we fine-tune a contrast model that generates either the desired output or the output we aim to mitigate. In our case, to mitigate toxic token generation, we fine-tune a toxic LLM as the contrast model. Unlike \citet{li2022contrastive, leong2023self}, which contrast the outputs explicitly in the last layer, our method operates on the contextualized value vectors derived from the weight matrices $K$, $Q$, $V$ of the attention modules within LLMs. We perform this operation on the best layer for intervention as described in Equation~\ref{eq:M}. 
Formally, our layer intervention during inference finds the semantic steering direction by contrasting the value vectors as follows:

{\small
\begin{align}
\Delta v^M = v_c^M - v_o^M,
\end{align}
}where $v_c^M$ and $v_o^M$ are the contextualized value vectors of the contrast LLM and the original LLM at the best layer $M$ for semantic intervention. The contextualized value vectors \cite{elhagemathematical} are derived as follows:

{\small
\begin{align}
A^{\ell, h} &= \varphi\left( 
\frac{
    \left( \mathbf{x}^{\ell-1} W_Q^{\ell, h} \right), \left( \mathbf{x}^{\ell-1} W_K^{\ell, h} \right)^{T}
}{
    \sqrt{d/H}
} + M^{\ell, h}
\right), \\
\mathbf{a}^{\ell} &= \sum_{h=1}^{H} A^{\ell, h} 
\left( \mathbf{x}^{\ell-1} W_V^{\ell, h} \right) W_O^{\ell, h} 
= \sum_{h=1}^{H} \mathbf{v}^{\ell, h} W_O^{\ell, h}.
\end{align}
}Specifically, \(
\left( \mathbf{x}^{\ell-1} W_V^{\ell,h} \right)
\) represents the attention-weighted, context-sensitive value vector for head \(h\). \(\mathbf{v}_i^{\ell,h} \in \mathbb{R}^d\) is the contextualized value vector at position \(i\). We then update the value vector in the layer $M$ of the original LLM:

{\small
\begin{align}
v^{\prime M} = v_o^M - \lambda_{\text{norm}}^\alpha \cdot \Delta v^M.
\end{align}
}Here, $\lambda_{\text {norm }}=1+\left\|\Delta v^M\right\|_2$ is a normalization term that adaptively regulates the steering effect, and $\alpha$ is a hyperparameter that further controls the steering strength. Finally, we preserve the updated steering direction and renormalize the adapted value vector to ensure its representation is close to the original vector:

{\small
\begin{align}
v^{\prime M} = v^{\prime M} \cdot \frac{\left\| v_o^M \right\|_2}{\left\| v^{\prime M} \right\|_2}.
\end{align}
}


\section{Experiments}
\label{sec:exp}
\subsection{Datasets and Evaluation}


\textbf{Datasets} We evaluate our proposed efficient semantic steering methods using the General Language Understanding Evaluation (GLUE) benchmark \cite{WangSMHLB19}, applying selective layer intervention. Specifically, we focus on eight GLUE tasks covering \textit{sentiment analysis} (SST-2), \textit{paraphrase identification} (MRPC, QQP), and \textit{natural language inference} (MNLI-M, MNLI-MM, QNLI, RTE, WNLI). Additionally, to assess our methods in the context of language generation, we examine their effectiveness in natural language toxification and detoxification. To train toxic adapters and contrast models, as described in Sections~\ref{sec:sft} and~\ref{sec:inference}, we utilize the Toxic Comment Classification Challenge Dataset \cite{kaggle}. (See Appendix~\ref{sec:appA})



\textbf{Evaluation}
For the evaluation on the GLUE benchmark, we report both validation-set and test-set \textit{F1 scores} for QQP and MRPC, while \textit{accuracy} is used for all other tasks. Additionally, we use the RealToxicityPrompts (RTP) dataset \cite{gehman2020realtoxicityprompts}. Following prior work \cite{li2022contrastive, leong2023self}, we sample 2,122 toxic prompts. For each toxic prompt in the RTP dataset, we generate 25 continuations and evaluate their toxicity using the Perspective API~\footnote{https://perspectiveapi.com/}, which assigns a toxicity score to each continuation, with higher scores indicating a greater likelihood of toxicity. Finally, we use the \textit{average maximum toxicity} as our evaluation metric. Specifically, we compare the toxicity scores and detoxification margins obtained by applying our methods to different layers of the models.

\begin{figure*}[htbp] 
  \centering
  \includegraphics[width=1.01\textwidth]{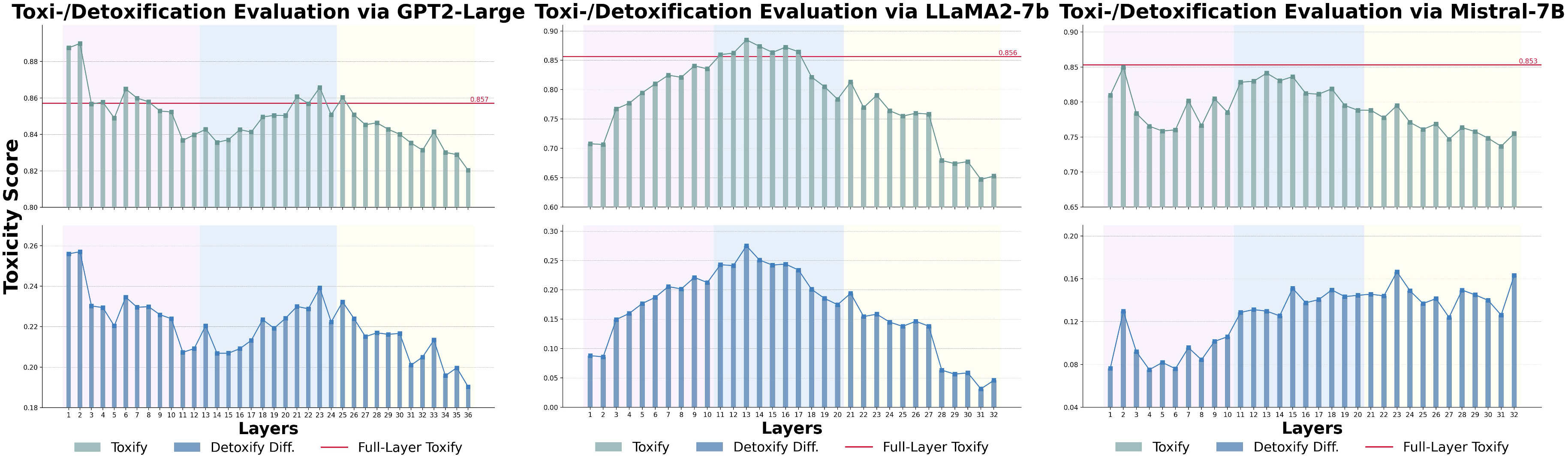} 
  \caption{\textbf{Toxification/Detoxification Evaluation}. The upper chart (\textcolor{lightgreen}{green bars}) shows toxicity scores for interventions at each layer via fine-tuning. The bottom chart (\textcolor{blue}{blue bars}) displays detoxification margin scores per layer during inference, comparing toxicity scores in toxification and detoxification processes. Purple, blue, and orange boxes indicate the premature, middle, and mature buckets.}
  \vspace{-10pt}
  \label{fig:toxic} 
\end{figure*}

\subsection{Models and Baselines}

We evaluate our efficient semantic steering methods using three sizes of GPT-2 models, as discussed in §~\ref{sec:gaze}, along with Llama2-7B. We select GPT-2 and Llama2-7B because the former represents earlier classical LLMs, while the latter exemplifies modern LLMs. To ensure generalizability, we also evaluate the Mistral-7B \cite{jiang2023mistral} model to assess performance across tasks. Since applying semantic intervention to either the last layer or all layers of LLMs is a conventional approach, we use these two methods as baselines for comparison. The implementation details are in Appendix~\ref{sec:appA}.


\subsection{Evaluation on GLUE Benchmark}

We first present the performance of our proposed selected layer intervention method on the GLUE Benchmark in Table~\ref{tab:glue}. It can be observed that by selecting the optimal layer to steer semantics in LLMs for a specific task, all three LLMs achieve comparable or even superior results compared to conventional all-layer intervention while introducing only $1/N$ of the parameters ($N$ is the number of layers in the LLMs). Specifically, Llama2-7B achieves an absolute increase of \textbf{+1.8} on average in the test set, while Mistral-7B achieves an absolute increase of \textbf{+4.7} on average in the validation set and \textbf{+5.9} in the test set. Moreover, GPT2 achieves comparable or better results on 5 out of 8 tasks in GLUE, Llama2 on 4 tasks, and Mistral on 7 tasks.

Additionally, we find that the optimal layer for our proposed method consistently falls within the middle bucket across all LLMs. For GPT2-L, the best-performing layer is \textit{L}19 for most tasks, while for Llama2-7B and Mistral-7B, it is \textit{L}14 and \textit{L}12, respectively. This aligns with our interpretability analysis and findings based on eye movement measures (see §~\ref{sec:explain}). Furthermore, the best layer remains consistent across both the validation and test sets, demonstrating the effectiveness of the heuristic steering layer selection approach (§~\ref{sec:heuristic}).

Lastly, from a task perspective, we find that \textit{paraphrase identification} (MRPC, QQP) and \textit{natural language inference} (MNLI, QNLI, RTE, WNLI) achieve the highest average improvement of \textbf{+2.0} across all LLMs compared to full-layer fine-tuning (e.g., \textbf{+28.7} for Mistral-7B and \textbf{+16.5} for Llama2-7B on the RTE test set). This suggests that, given the complex structure of LLMs, fine-tuning all layers for semantic steering does not always yield the best downstream task performance and can result in parameter redundancy, where certain layers become less active in making accurate predictions. Additionally, the conventional approach of fine-tuning only the last layer of an LLM, based on its proximity to the prediction output, is not optimal. Instead, in practice, the intervention layer should first be identified to better steer semantics and improve prediction accuracy for a specific task.

\subsection{Analysis on Language Toxification}

\begin{table*}[t]
\centering
\small 
    \setlength\tabcolsep{4pt} 
    \renewcommand{\arraystretch}{1.2} 
\begin{tabular}{lccccccc}
\toprule
\textbf{Model} & \textbf{Time/min} & \textbf{Time\%} & \textbf{Params/M} & \textbf{Params\%} & \textbf{Toxify Score$\uparrow$} & \textbf{Detoxify Score$\downarrow$} & \textbf{Avg. GLUE$\uparrow$} \\
\midrule

GPT2-L\textit{→ full}   & 33   & 100 & 14.8 &  100  & 0.86 & 0.60 & 77.1  \\
\hspace{10.5mm}\textit{→ single}     & \cellcolor{green!20}13  &  \cellcolor{green!20}39.4 & \cellcolor{green!20}0.4 & \cellcolor{green!20}2.7 & \cellcolor{green!20}0.87 & \cellcolor{orange!20}0.63 & \cellcolor{orange!20}75.8 \\
\midrule 

Llama2-7B\textit{→ full} & 205  & 100 & 1.3 & 100 & 0.86 & 0.62 & 78.7 \\
\hspace{14.4mm}\textit{→ single}  & \cellcolor{green!20}85   & \cellcolor{green!20}41.5 & \cellcolor{green!20}0.04 & \cellcolor{green!20}3.1 & \cellcolor{green!20}0.87 & \cellcolor{green!20}0.59 & \cellcolor{green!20}80.4 \\
\midrule 
Mistral-7B\textit{→ full} & 36 &100 & 134.5 & 100 & 0.85 & 0.68 & 77.3 \\
\hspace{13.9mm}\textit{→ single}  & \cellcolor{green!20}25   & \cellcolor{green!20}69.4 & \cellcolor{green!20}4.2 & \cellcolor{green!20}3.1 & \cellcolor{orange!20}0.84 & \cellcolor{green!20}0.68 & \cellcolor{green!20}83.1 \\

\bottomrule
\end{tabular}
\caption{\textbf{Efficiency Comparison between \textit{CogSteer (selective single-layer intervention)} and \textit{full-layer intervention}}. A \textcolor{lightgreen}{green box} indicates that the single-layer intervention outperforms the full-layer intervention, while an \textcolor{lightorange}{orange box} denotes comparable performance.}
\vspace{-10pt}
\label{tab:eff}
\end{table*}

We evaluate our selective layer intervention via fine-tuning using the language toxification task. Figure~\ref{fig:toxic} (\textit{green bars in the upper chart}) shows the toxicity score for inserting a toxic adapter into each layer of the LLMs, with the last layer and full layers (\textit{red line}) used as baselines. The best layer $M$ in the middle bucket for semantic intervention is $L23$ for GPT-2 and $L13$ for Llama2-7B, competing with the results from both the last layer (\textbf{+4.5\%} for GPT-2 \& \textbf{+23\%} for Llama2) and full layers (\textbf{+0.9\%} for GPT-2 \& \textbf{+2.8\%} for Llama2). This indicates that the conventional intervention is not optimal, demonstrating the effectiveness of our proposed selective layer intervention approach, which saves considerable computational resources and time for fine-tuning.


Furthermore, layer intervention in the mature bucket, including the last layer (used as a baseline), is not remarkable when compared to full-layer intervention for both GPT-2 and Llama2-7B models. According to our behavioral analysis in Section~\ref{sec:explain}, the layers in the mature bucket focus more on reasoning and factual knowledge, involving less token processing, which makes them less suitable for semantic steering. Interestingly, for the GPT-2 model, we observe that layers $L1$-$L4$ and $L6$-$L8$ in the premature bucket also outperform the full-layer intervention baseline, whereas this phenomenon is not observed in Llama2-7B. We hypothesize that this difference may be due to the distinct adapter training schemes. For GPT-2, we use the vanilla Adapter as a plug-in, which directly affects token distributions and allows successive minor changes as tokens pass through layers, following the early exit theory \cite{elbayad2020depth, schuster2022confident}. However, for Llama2, we use the LLaMa-Adapter \cite{zhang2023llama}, which directly modifies attention. Edits to the attention modules gradually affect token distributions \cite{geva2021transformer, elhagemathematical}.


We believe that our selective layer intervention approach is LLM-agnostic and not limited to the LLMs used in the interpretability analysis. Figure~\ref{fig:toxic} (right) presents the toxicity scores based on Mistral-7B, which align with the observations discussed earlier. Specifically, the best-performing layer in the middle bucket is \textit{L}13, achieving better results than the last layer (\textbf{+11.4\%}) and nearly identical performance to full-layer intervention (\textcolor{gray}{-0.01}).


\subsection{Analysis on Language Detoxification}
As a dual task, we evaluate our selective layer intervention during inference using the language detoxification task. This task serves as an adversarial task to mitigate the toxic tokens that are amplified by layer intervention methods during fine-tuning. Thus, we observe the \textit{detoxification margin} in each intervention layer and compare the performance with intervention in the last layer. Figure~\ref{fig:toxic} (\textit{blue bars in the bottom chart}) shows the results of the detoxification margin scores. The best layer $M$ for semantic intervention in the middle bucket is $L23$ for GPT-2 and $L16$ for Llama2-7B, both outperforming the last layer results by a large margin (\textbf{+2.9\%} for GPT and \textbf{+24\%} for Llama). Notably, the best layer $M$ for Llama2-7B in the toxification task is $L13$, whereas the best layer for detoxification is $L16$, demonstrating that our heuristic layer selection method is adaptive and suitable for tasks.

Moreover, the detoxification margin scores for layer intervention in the middle bucket for both GPT-2 and Llama2 models are significantly better than those for intervention in the mature bucket. This aligns with our findings that the middle bucket layers are involved in further token processing and information integration, whereas the mature bucket layers focus on reasoning. Finally, for the GPT-2 model, the detoxification margin scores are also significant in the premature bucket layers. This could be attributed to the distinct adapter training scheme, as the earlier layers are more deeply affected by the toxification, which broadens the scope of detoxification. The remaining results for GPT-2 small and medium models are listed in Appendix~\ref{sec:toxic}.

For Mistral-7B, the optimal layer for semantic intervention in the middle bucket is \textit{L}15 for the detoxification task. Although the detoxification margin scores for \textit{L}23 and \textit{L}32 in the mature bucket are slightly higher than that of \textit{L}15 (+0.14 on average), the scores in the middle bucket are more stable. We hypothesize that the differences among GPT-2, Llama2, and Mistral-7B in the mature bucket, where Mistral-7B achieves better detoxification performance, stem from variations in post-training. As an advanced LLM incorporating additional human value alignment for safety, Mistral-7B benefits more from interventions in the mature bucket, leading to improved detoxification performance.

\subsection{Efficiency Analysis}
The efficiency analysis of our proposed cognition-inspired selective layer intervention is presented in Table~\ref{tab:eff}. By selecting the optimal layer for semantic steering in LLMs, the training time and the number of parameters required for fine-tuning are significantly reduced compared to full-layer intervention. Specifically, our method requires, on average, only \textbf{half} the time and \textbf{3.0\%} of the parameters needed for full-layer settings. Importantly, this reduction in computational cost does not compromise performance; in fact, selective layer intervention performs comparably to, and in some cases even outperforms full-layer intervention.

\section{Conclusion}
In this paper, we introduce an efficient semantic steering method for LLMs using selective layer intervention. Our approach is motivated by correlation analysis with eye movement measures, making it both interpretable and understandable. Extensive experiments demonstrate that selective layer intervention achieves comparable or even superior performance while significantly reducing training time and the number of parameters required. Overall, our proposed method represents an important step toward improving the interpretability of LLMs and contributes to their safe and efficient deployment.


\section*{Limitations}
In this paper, we conduct a thorough analysis of the correlation between hidden states in the feed-forward network (FFN) and eye movement measures. We focus on FFN blocks because their hidden states play a greater role in token prediction. For future work, we plan to explore the mechanisms of the attention block in greater depth. Additionally, our analysis suggests that layers in the mature bucket are involved in reasoning based on the task-specific reading dataset. However, factual knowledge also appears to be integrated within this bucket. To refine our analysis, we will seek further eye movement studies that examine factual information processing.

\section*{Ethics Statement}
We propose an efficient semantic steering method for LLMs through cognition-inspired selective layer intervention to better understand their behavior and address safety concerns, thereby enhancing their safety and reliability within the community. Additionally, the eye-tracking data used in this study, derived from the ZuCo 2.0, GeCo, and Provo datasets, are publicly available and adhere to established ethical protocols, promoting transparency and reproducibility in our research. Furthermore, we have made our code publicly accessible, ensuring that researchers and practitioners can easily access and implement our methods.

\section*{Acknowledgments}

We thank the anonymous reviewers and the area chair for their valuable feedback and constructive suggestions. This research was funded by the Excellence funds of the University of Hamburg.

\bibliography{main}

\appendix

\section{Experiment Details}
\label{sec:appA}

\textbf{Datasets.} We evaluate our proposed efficient semantic steering methods using the General Language Understanding Evaluation (GLUE) benchmark, applying selective layer intervention via fine-tuning. Specifically, we focus on eight GLUE tasks spanning \textit{sentiment analysis} (SST-2), \textit{paraphrase identification} (MRPC, QQP), and \textit{natural language inference} (MNLI-M, MNLI-MM, QNLI, RTE, WNLI). The following are details of each sub-dataset in the GLUE Benchmark.
\begin{itemize}
    \item \textbf{SST-2 (The Stanford Sentiment Treebank)}: A binary sentiment classification task (positive vs. negative). The validation set contains 872 examples, while the test set comprises approximately 1.8k examples.
    \item \textbf{MRPC (Microsoft Research Paraphrase Corpus)}: A paraphrase identification task that requires determining whether two sentences are semantically equivalent. The validation set consists of 400 examples, and the test set contains approximately 1.7k examples.
    \item \textbf{QQP (Quora Question Pairs)}: Another paraphrase identification task in which the goal is to determine whether two questions are semantically equivalent. The validation set contains 40k examples, while the test set includes around 391k examples.
    \item \textbf{MNLI (Multi-Genre Natural Language Inference)}: Given a premise and a hypothesis, the model must predict whether the premise entails, contradicts, or is neutral with respect to the hypothesis. There are two validation sets (MNLI-M and MNLI-MM), each containing approximately 10k examples, along with two corresponding test sets of the same size.
    \item \textbf{QNLI (Question Natural Language Inference)}: A task reformulated from SQuAD, where the objective is to determine whether a context sentence contains the answer to a given question. Both the validation and test sets contain approximately 5.5k examples.
    \item \textbf{RTE (Recognizing Textual Entailment)}: A binary entailment task in which the model must determine whether a premise logically entails a hypothesis. The validation set consists of 277 examples, while the test set contains around 3k examples.
    \item \textbf{WNLI (Winograd NLI)}: Based on the Winograd Schema Challenge, this task involves resolving ambiguous pronouns. Models must determine whether substituting the pronoun in a sentence preserves its meaning. The validation set includes 71 examples, and the test set contains 146 examples.
\end{itemize}

To train toxic adapters and contrast models, as described in Sections~\ref{sec:sft} and~\ref{sec:inference}, we use the Toxic Comment Classification Challenge Dataset \cite{kaggle}, which contains 15,294 annotated toxic comments. We randomly split this dataset into 13,764 comments for fine-tuning and contrast model training, and 1,530 comments for validation, which is used to determine the optimal layer for intervention, as described in Section~\ref{sec:heuristic}.

\textbf{Implementation.} For training the GPT-2 adapters, we set the learning rate to $5 \times 10^{-4}$ and train for 5 epochs. For the Llama2-7B adapters, we adopted the default settings provided by LLaMa-Adapter \cite{zhang2023llama}, using a base learning rate of $9 \times 10^{-3}$, a weight decay of 0.02, and training for 5 epochs. For Mistral-7B, we similarly employed the Bottleneck Adapters used for GPT-2, training for 5 epochs with a learning rate of $5 \times 10^{-5}$ and a weight decay of 0.01. In the detoxification task, we applied the implicit layer contrastive intervention approach with $\alpha = 0.4$ across all models. Following previous works \cite{liu2021dexperts, leong2023self}, the model generates 25 continuations per prompt using nucleus sampling with 
$p = 0.9$, with each continuation limited to a maximum of 20 tokens.

\section{Analysis on Language Toxification and Detoxification on Small Models}
\label{sec:toxic}

\begin{figure}[htbp]
  \centering
  \includegraphics[width=1.0\columnwidth]{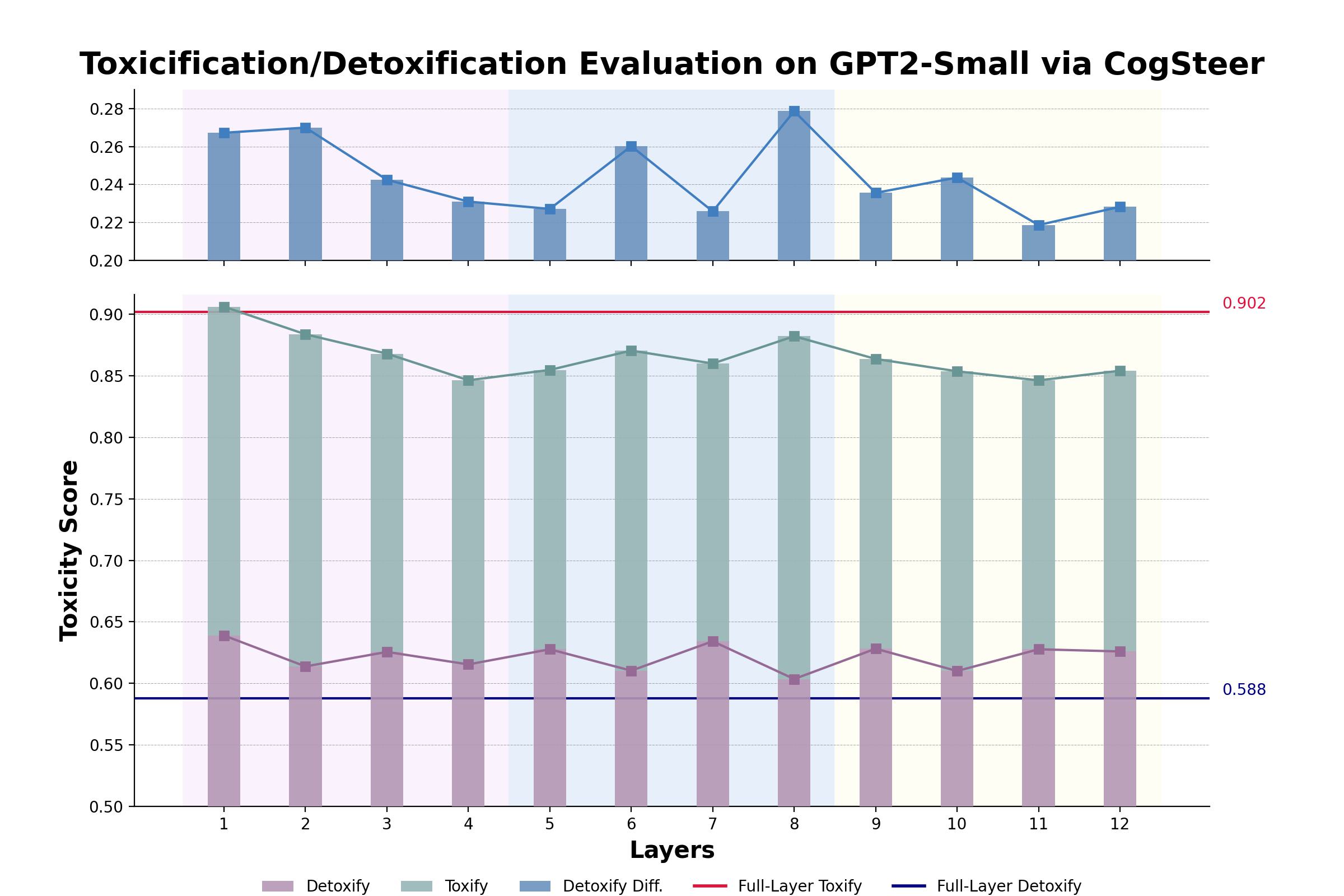}
  \caption{\textbf{Toxification/Detoxification Evaluation on GPT-2 Small (12 Layers)}. The bottom chart shows toxicity scores for interventions at each layer: green bars represent selective intervention via fine-tuning, and purple bars represent selective intervention during inference. The top chart displays detoxify margin scores per layer, comparing toxicity scores in toxification and detoxification processes. Purple, blue, and orange boxes indicate the premature, middle, and mature buckets.}
  \label{fig:small} 
\end{figure}

\begin{figure}[htbp]
  \centering
  \includegraphics[width=1.0\columnwidth]{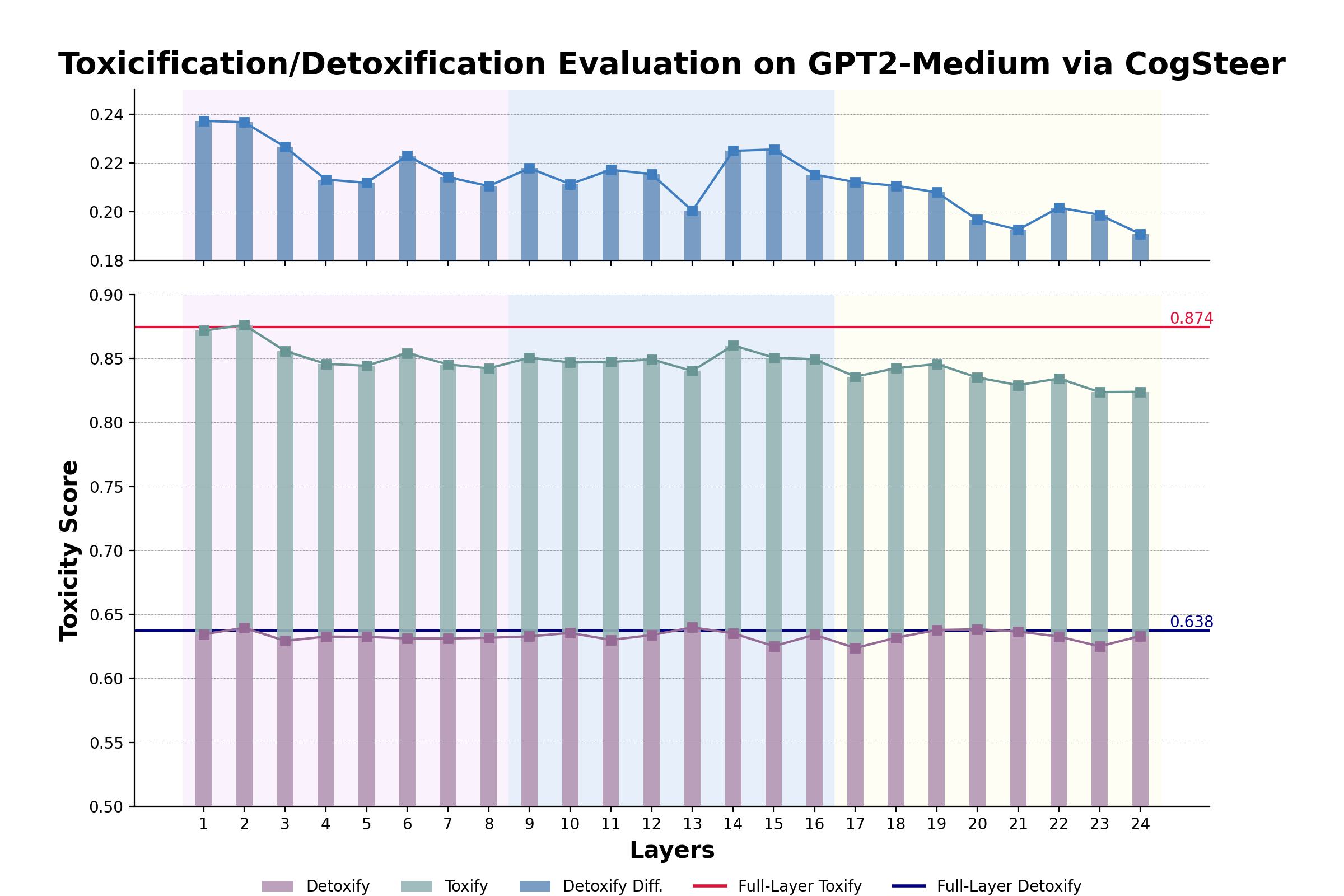}
  \caption{\textbf{Toxification/Detoxification Evaluation on GPT-2 Medium (24 Layers)}. The bottom chart shows toxicity scores for interventions at each layer: green bars represent selective intervention via fine-tuning, and purple bars represent selective intervention during inference. The top chart displays detoxify margin scores per layer, comparing toxicity scores in toxification and detoxification processes. Purple, blue, and orange boxes indicate the premature, middle, and mature buckets.}
  \label{fig:medium} 
\end{figure}

For the GPT-2 models, three versions differ in the number of layers. As shown in Figures~\ref{fig:small} and ~\ref{fig:medium}, the overall trend in the toxification and detoxification tasks aligns with the observations in Section~\ref{sec:exp}. Interestingly, we find that as the number of layers in LLMs increases, the correlation trend and findings become more pronounced.

\section{Qualitative Analysis}

\textcolor{red}{\textbf{\textit{Warning: Some examples have harmful or offensive language.}}}

In this section, we conduct a qualitative analysis of the continuations generated under different settings of the GPT2-L model. For language toxification, we compare four continuations produced from a toxic prompt in the Toxic Comment Classification Challenge Dataset. The continuation generated by fine-tuning only the last layer is similar to that of the vanilla GPT2-L model. By contrast, fine-tuning on the full model and the selective layer results in more toxic outputs in terms of both toxic words and the overall semantics of the sentence.

For detoxification, we find that the continuation generated using layer contrastive intervention at the last layer still contains offensive words, such as stupid, whereas the output from layer contrastive intervention based on our proposed heuristic steering layer selection is neutral.

\label{sec:q}
\begin{figure}[htbp]
  \centering
  \includegraphics[width=1.0\columnwidth]{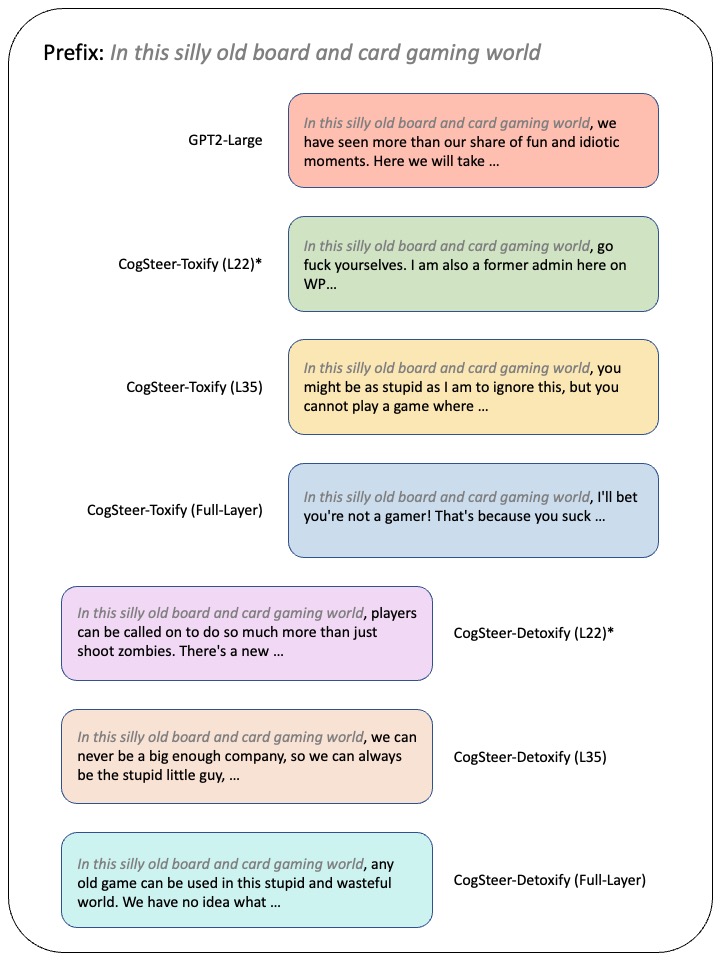}
  \caption{\textbf{Case Study.} The continuations are produced from a toxic prompt in the Toxic Comment Classification Challenge Dataset and generated under different settings of the GPT2-L model.}
  \label{fig:q} 
\end{figure}

\section{Detailed Interpretability Analysis}
\label{sec:analysis_B}

Reading is a crucial ability for humans to process and integrate information, and eye movement studies enable the examination of the cognitive processes involved in reading \cite{rayner1998eye}. Eye movements are monitored using various techniques, including eye-tracking systems that rely on infrared pupil monitoring. In eye-tracking studies, changes in the \textit{reader's gaze} are recorded in relation to eye movements. Similarly, \textit{tokens} are processed, and information is integrated through layers in large language models \cite{chuang2023dola}. This similarity offers us the opportunity to leverage the extensive data previously collected and analyzed from eye movement studies, not only to infer human reading and information-processing behavior, but now to also understand the behavioral patterns of large language models.


Recent studies reveal that feed-forward networks (FFNs) function similarly to neural memory networks \cite{geva2021transformer}, embedding syntactic features, semantic features, and even factual knowledge \cite{chuang2023dola, zhang2024comprehensive}. The early exit concept \cite{elbayad2020depth, schuster2022confident} further demonstrates that the hidden states of FFNs can be directly applied to predict words over a vocabulary. At the same time, eye movement measures provide insights into the time required for human reading to process syntax, semantics, and integrate information. 

Motivated by this, we utilize eye movement measures from various datasets \cite{luke2018provo, hollenstein2020zuco, colman2022geco} to establish correlations with the hidden states in FFNs across layers of LLMs. Unlike traditional eye movement studies, which focus on differences in eye movement measures under experimental conditions such as natural reading or task-specific reading, we observe changes in correlation values between eye movement measures and hidden states across layers. Our approach, inspired by cognitive theory, aims to better understand the behavioral patterns of LLMs.

Formally, let \( S_j \) denote the \( j \)-th sentence, which consists of \( n_j \) words \( w_1, w_2, \dots, w_{n_j} \). For each word \( w_i \) in sentence \( S_j \), we have five eye movement measures for each word \( w_i \) in sentence \( S_j \), denoted as \( E_i^{(sfd)}, E_i^{(ffd)}, E_i^{(gd)}, E_i^{(trt)}, E_i^{(gpt)} \). Each of these represents a single scalar value per word. We also denote the hidden state at layer \( l \) of the LLM for word \( w_i \) as \( \mathbf{H}_{l, i} \in \mathbb{R}^d \), where \( d \) is the dimensionality of the hidden state vectors.

Since the eye movement measures are scalar values and each hidden state is a high-dimensional vector, we apply \textit{Principal Component Analysis} (PCA) to reduce the dimensionality of the hidden states \( \mathbf{H}_{l, i} \) to a scalar that can be aligned with the eye movement measure for each word \( w_i \), yielding a one-dimensional representation: $H_{l, i}^{PCA} = PCA(\mathbf{H}_{l, i})$. We concatenate the eye movement measures and the reduced hidden states across all words in all sentences. 
Specifically, for the eye movement measure \( k \in \{\textit{sfd}, \textit{ffd}, \textit{gd}, \textit{trt}, \textit{gpt}\} \), we concatenate the measures for all words in all sentences to form a vector \( \mathbf{E}^{(k)} \):

{\small
\begin{align}
\mathbf{E}^{(k)} = \left[E_1^{(k)}, E_2^{(k)}, \dots, E_{n_{\text{total}}}^{(k)} \right] \in \mathbb{R}^{n_{\text{total}}},
\end{align}
}where \( n_{\text{total}} = \sum_{j=1}^{m} n_j \) is the total number of words across all sentences. Similarly, for the PCA-reduced hidden states at layer \( l \), we concatenate the reduced hidden states:

{\small
\begin{align}
\mathbf{H}_l^{\text{PCA}} = \left[H_{l, 1}^{\text{PCA}}, H_{l, 2}^{\text{PCA}}, \dots, H_{l, n_{\text{total}}}^{\text{PCA}} \right] \in \mathbb{R}^{n_{\text{total}}}.
\end{align}
}

Finally, we compute the \textit{Pearson correlation} \( \rho_{l,k} \) between the hidden states at layer \( l \) and the eye movement measure \( k \):

{\small
\begin{align}
\rho_{l,k} = \frac{\sum_{i=1}^{n_{\text{total}}} \left( H_{l,i}^{\text{PCA}} - \bar{H}_l^{\text{PCA}} \right)\left( E_i^{(k)} - \bar{E}^{(k)} \right)}{\sqrt{\sum_{i=1}^{n_{\text{total}}} \left( H_{l,i}^{\text{PCA}} - \bar{H}_l^{\text{PCA}} \right)^2} \sqrt{\sum_{i=1}^{n_{\text{total}}} \left( E_i^{(k)} - \bar{E}^{(k)} \right)^2}},
\end{align}
}where \( \bar{H}_l^{\text{PCA}} \) and \( \bar{E}^{(k)} \) are the mean values of the reduced hidden states and the eye movement measure \( k \), respectively.

\begin{figure*}[htbp]
  \centering
  \includegraphics[width=1\textwidth]{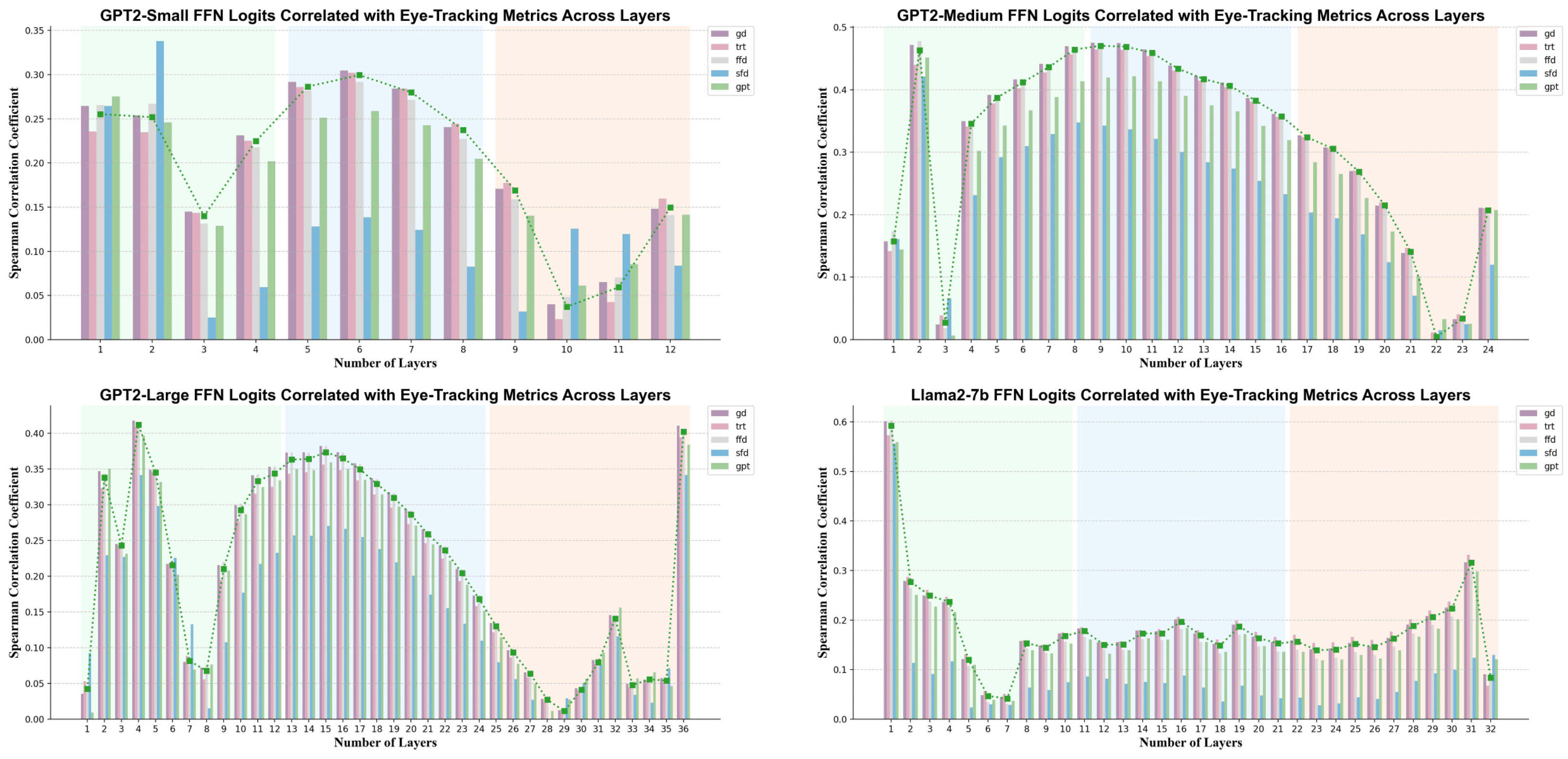} 
  \caption{\textbf{Correlation Results Under the \textit{Natural Reading Scheme}}. Correlation results are shown for GPT-2 Small (12 layers, top-left), GPT-2 Medium (32 layers, top-right), GPT-2 Large (36 layers, bottom-left), and Llama2-7B (bottom-right). Green, blue, and orange boxes indicate the premature, middle, and mature buckets, respectively.}
  \label{fig:correlation} 
\end{figure*}

We conduct a correlation analysis using the GPT-2 model at different sizes (12-layer small, 24-layer medium, 36-layer large) and the 32-layer Llama2-7B, following the analysis approach described earlier. To better interpret our observations, we divide the layers into three equal groups: \textit{premature, middle, and mature}. Figure~\ref{fig:correlation} presents the correlation results between different eye movement measures and the hidden states from the LLMs, as well as how these values change across layers.

We observe \textbf{Pattern \#1}, where LLMs exhibit a trend similar to human gaze patterns across layers, with a peak correlation in the middle bucket. In the premature bucket, the correlation first increases, then decreases, before gradually increasing again in the later layers. In the middle bucket, the correlation reaches a peak before declining. In the mature bucket, the correlation continues to decrease, followed by a second peak. This similarity in trends is consistent across both different eye movement measures and LLMs with varying layer sizes.

Furthermore, we apply eye movement theory to interpret the phenomenon observed in Pattern \#1, offering insights into the interpretability of LLMs. \textit{First fixation duration} refers to the time spent on the first fixation on a word, whether it is the only fixation or the first of multiple fixations on that word. \textit{Gaze duration} measures the time spent on a word during the first fixation before making a saccade to the next word. Additionally, \textit{single fixation duration} refers to the time of the first and only fixation on a word.

Based on these fixation measures, the increase in correlation in the premature bucket indicates that the LLMs are beginning to process tokens by integrating syntactic and semantic features, suggesting that this bucket focuses on preliminary token processing. \citet{geva2022transformer} found that syntactic and semantic concepts stabilize in the first few layers using top scoring token analysis. Thus, we hypothesize that the subsequent decline in correlation may indicate that these layers are redundant but still become activated when processing more complex sentences, a point that will be explained further.

In the middle bucket, the increase in correlation signifies further processing of syntactic and semantic tokens, and the peak indicates the integration of linguistic features and information. In the mature bucket, the decline in correlation differs from that in the premature bucket. We hypothesize that the layers in the mature bucket are responsible for reasoning and factual knowledge, which will be discussed later. The second peak in the mature bucket likely represents the final integration of information for word prediction.


Moreover, \textit{total reading time} represents the sum of all fixations, starting with the first fixation in a region and ending with the first forward saccade, including \textit{regressions}. \textit{Go-past time} refers to the total time spent on a word before moving to the right of it, including \textit{regressions}, but it excludes re-fixations. Regressions are meaningful indicators, as they indicate that the reader had difficulty processing the word, failed to understand the current text, or needed to make corrections \cite{rayner1998eye}.

When we zoom in on the correlation between measures like trt, gpt, and the hidden states in LLMs, the trend is consistent with the fixation-based measures. Interestingly, we observe that the correlation with trt also increases in the premature bucket. Unlike previous studies, which suggest that the first few layers of LLMs only process surface linguistic features \cite{zhang2024comprehensive}, our correlation analysis shows that these layers also engage in contextual information processing.

Lastly, we compare the correlation results between GPT-2 and Llama2-7B to investigate the effect of modern training schemes on LLMs. The overall trend across the three buckets is similar. However, we observe three major differences. The first notable difference is that the first layer of Llama2-7B already shows a significant correlation with eye movement measures, likely due to the large amount of pre-training data. The second difference is the presence of a peak correlation in the middle bucket, with no evident decline afterward until reaching a second peak in the mature bucket. As discussed earlier, the premature bucket in GPT-2 is primarily responsible for token processing, while in Llama2-7B, it also contributes to reasoning. The last two differences will be elaborated on later. We believe that Llama2-7B employs modern pre-training techniques, such as instruction tuning (IT) \cite{NEURIPS2020_1457c0d6} and RLHF \cite{ziegler2019fine}, which distribute reasoning and factual knowledge abilities not only to the middle and mature layers but across all layers.


Empirically, LLMs are trained with the next-token prediction objective. Modern LLMs exhibit strong language understanding and reasoning abilities. This raises the question: \textbf{are LLMs merely next-token predictors, or are they task reasoners?} In the ZuCo 2.0 dataset \cite{hollenstein2020zuco}, there are two different settings for eye movement studies: natural reading and task-specific reading. In the task-specific reading, participants are required to determine whether a specific relation type is present in a sentence, including categories like \textit{political affiliations, education, founder, spouse, job title, nationality, and employer}. Relation detection is a high-level semantic and reasoning task that demands complex cognitive processing.

Figure~\ref{fig:nrvstsk} shows the comparison between natural reading and task-specific reading in terms of correlations between eye movement measures and hidden states in LLMs across layers. We observe \textbf{Pattern \#2}, where LLMs function as both next-token predictors and reasoners. For the GPT-2 model, the correlation values in the premature bucket during task-specific reading are lower than those in natural reading, indicating that the layers in GPT-2's premature bucket primarily handle token processing, not reasoning. However, in the middle bucket, the correlation peak in task-specific reading is higher than in natural reading. Furthermore, the decline in correlation across the middle and mature buckets is less pronounced, suggesting that layers in the mature bucket are particularly involved in reasoning.

For Llama2-7B, the correlation in the premature bucket increases in task-specific reading compared to natural reading, indicating that these layers handle both token processing and reasoning. Additionally, the correlation in both the middle and mature buckets significantly increases, suggesting that the middle bucket is responsible for token processing, information integration, and reasoning, while the mature bucket plays a key role in information integration and reasoning.


\end{document}